\documentclass{article}
\usepackage{iclr2027_conference,times}
\usepackage[T1]{fontenc}
\usepackage[utf8]{inputenc}
\usepackage{amsmath,amssymb,amsfonts,bm}
\usepackage{graphicx}
\usepackage{booktabs}
\usepackage{longtable}
\usepackage{array}
\usepackage{calc}
\usepackage{multirow}
\usepackage{xcolor}
\usepackage{url}
\usepackage{xurl}
\usepackage{hyperref}
\usepackage{microtype}
\usepackage{enumitem}
\hypersetup{hidelinks}

\title{When Does Action Credit Need Updating?}

\author{
Hongye Yang \\
College of Computing\\ Georgia Institute of Technology \\
\texttt{hyang783@gatech.edu}
\And
Boxiao Huang \\
College of Computing\\ Georgia Institute of Technology \\
\texttt{bhuang361@gatech.edu}
}

\iclrfinalcopy

\newcommand{\abs}[1]{\left|#1\right|}

\begin{document}
\maketitle
\lhead{Preprint}
\maketitle

\begin{abstract}
Tool-using agents are continually updated with new interaction data. After each policy update, however, previously estimated action credits may become stale. Recomputing them from scratch can require many additional tool calls and environment interactions, making repeated updates increasingly expensive. We ask a simple question: \emph{when does historical action credit actually need to be updated?} Our key observation is that a change in action value does not necessarily imply a change in the decision. Historical credit can still be useful as long as policy-induced drift is too small to overturn the existing action ranking. Building on this idea, we introduce pairwise branch sensitivity to capture how strongly a policy update affects the downstream regions that distinguish two candidate actions. We then derive a first-order anchored credit-transport estimator that updates historical credit using old interventional trajectories, and propose a Decision-Sufficient Credit Gate (DSC-Gate) that chooses whether to reuse, transport, or resample credit. Experiments show that branch sensitivity explains credit drift substantially better than global policy distance. With sufficient historical data, credit transport reduces estimation error, while its benefit to decision making is concentrated on updates that affect action-distinguishing branches. On a fully independent test set, DSC-Gate changes mean regret by only $+0.00004$ relative to a gap-based gate while reducing mean new tool steps from 472 to 286, a 39.4\% reduction. We observe the same pattern after a real tool-agent parameter update. Overall, our results show that agents do not need to recompute action credit after every policy update: much of the historical evidence can be reused or cheaply corrected, reducing the additional interaction required to keep action decisions up to date.
\end{abstract}

\section{Introduction}

Consider a retail support agent handling a refund request. Given the same user problem, it may first inspect the order status or first verify refund eligibility, before invoking downstream tools to complete the workflow. After many historical interactions, the agent has accumulated evidence about which route is more reliable. The policy is then updated, perhaps learning to recover from failed tool calls or handle exceptional orders more effectively. The downstream consequences of the original actions may therefore change. Must the agent now execute these routes again and recompute their credit from scratch?

Often, this is unnecessary. A policy update may improve downstream behavior shared by several action branches while leaving their relative ordering unchanged. Conversely, a much smaller update can matter immediately when it affects states reached primarily by one candidate action. Thus, neither global policy change nor numerical value drift alone determines whether historical credit has become unusable. What matters for the current decision is whether the induced change is large enough, and sufficiently aligned with the competing branches, to invalidate the existing action comparison.

This distinction is increasingly relevant for tool-using agents that improve through repeated interaction and tool-use training \citep{li2023apibank}. Stepwise evaluation has made the reliability of individual tool decisions increasingly measurable \citep{chen2024teval}, while large-scale benchmarks expose instability across tools and environments \citep{guo2024stabletoolbench}. Stateful evaluation further emphasizes that an early action is valuable only through the downstream behavior that follows it \citep{lu2025toolsandbox}. Once the downstream policy changes, action credit estimated under an earlier policy can therefore become stale. This dependence of return on the continuation policy is fundamental to credit assignment \citep{sutton1988learning}, and recent tool-agent studies also report unstable tool preferences across settings \citep{faghih2025toolpreferences}. Although off-policy evaluation provides principled ways to reuse historical trajectories under a changed policy \citep{dudik2014doublyrobust}, re-evaluating every candidate action from scratch remains the most direct way to avoid stale credit and repeatedly incurs additional tool calls, environment interaction, and sampling. We therefore ask a more targeted question: \emph{when does historical action credit actually need to be updated?}

Our starting point is to separate \emph{credit drift} from \emph{decision failure}. An action comparison can change numerically without changing the decision it supports. Whether this happens depends on both the existing action gap and where the policy update acts downstream. Updates concentrated in regions commonly visited by competing branches may largely cancel in their relative credit, whereas similarly sized updates aligned with branch differences can induce much larger drift. We capture this structure through \emph{pairwise branch sensitivity}, which measures how strongly a policy update overlaps with downstream regions that distinguish two candidate actions.

\begin{figure}[!htbp]
  \centering
  \includegraphics[width=\textwidth]{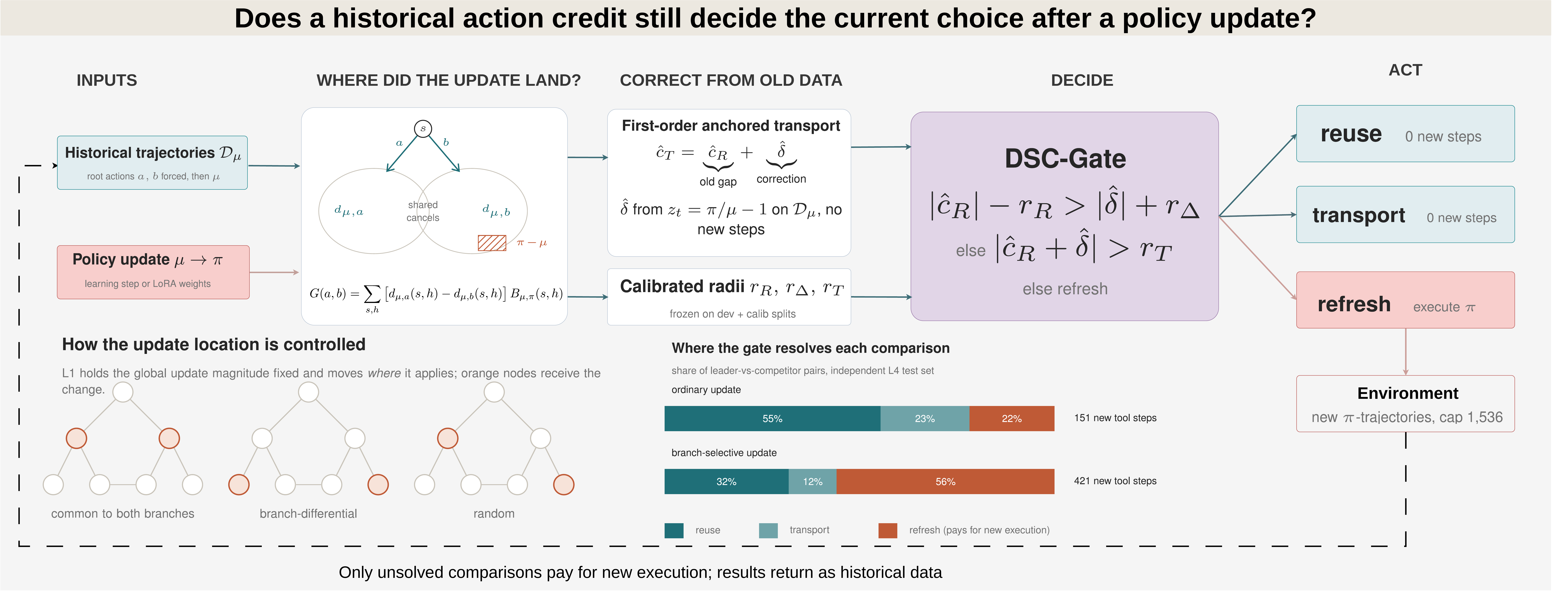}
    \caption{\textbf{Deciding when historical action credit must be refreshed.}
    Left to right: after a policy update $\mu\!\to\!\pi$, old interventional trajectories
    $\mathcal{D}_\mu$ collected under the previous policy are reused to compare two
    candidate root actions $a$ and $b$. Pairwise branch sensitivity $G(a,b)$ weights the
    update by the visitation difference $d_{\mu,a}-d_{\mu,b}$, so changes in states that
    both branches visit cancel in the action difference, and only changes in
    branch-differential states drive relative drift. First-order anchored transport then
    corrects the historical gap as $\widehat c_T=\widehat c_R+\widehat\delta$ from the same
    trajectories, without new execution. DSC-Gate combines $\widehat c_R$,
    $\widehat\delta$, and empirically calibrated radii $r_R,r_\Delta,r_T$ to resolve each
    leader-versus-competitor comparison by reuse, transport, or refresh; only refresh
    executes the target policy in the environment. The bars show how the gate shifts toward
    refresh when the update is concentrated on branch-differential states.}
  \label{fig:overview}
\end{figure}

Historical credit that has drifted also need not be discarded immediately. We derive a first-order anchored credit-transport estimator that uses historical interventional trajectories to estimate the policy-induced correction while retaining the old action comparison as an anchor. Building on this estimator, we introduce the \emph{Decision-Sufficient Credit Gate} (DSC-Gate), which determines whether a comparison can be resolved by directly reusing historical credit, transporting it from old trajectories, or collecting new target-policy data. New execution is therefore reserved for comparisons that cannot be resolved from existing evidence. Figure~\ref{fig:overview} summarizes this pipeline end to end

We evaluate this framework through four controlled levels followed by a separate real-tool validation. L1 tests whether branch sensitivity explains credit drift when global update magnitude is controlled. L2 evaluates first-order transport under finite historical data. L3 asks when improved credit estimation actually improves action selection. L4 freezes the resulting gate and evaluates it on a fully independent test population. We then test whether the same mechanism appears after an actual parameter update of a tool-using language model. Across these experiments, pairwise branch sensitivity explains credit drift substantially better than global policy distance, while transport reduces credit-estimation error when sufficient historical data are available. Its decision-level benefit is concentrated on updates that can alter action rankings. On the independent L4 test set, DSC-Gate changes mean regret by only $+0.00004$ relative to a gap-based gate while reducing mean new tool steps from 472 to 286, a 39.4\% reduction. The real model-weight update exhibits the same conditional pattern.

Our main contributions are:
\begin{itemize}
    \item We distinguish credit drift from decision failure after a policy update, showing why numerical changes in action value alone do not determine whether historical credit should be refreshed.
    
    \item We introduce pairwise branch sensitivity, a local quantity that characterizes how strongly a policy update affects downstream regions that distinguish candidate actions.
    
    \item We derive a first-order anchored credit-transport estimator that updates historical action comparisons using existing interventional trajectories before requiring new target-policy execution.
    
    \item We introduce DSC-Gate, which combines the historical action gap, transported correction, and empirical uncertainty to adaptively choose among reuse, transport, and resampling.
\end{itemize}

Taken together, these results show that maintaining action credit after a policy update need not require recomputing every candidate action from scratch. Historical evidence can often be reused, and when it becomes partially stale, existing trajectories can frequently provide a useful correction before new execution is needed. This reduces repeated tool calls, environment interaction, and resampling, lowering the marginal cost of keeping agent decisions up to date as their policies evolve.

\section{Related Work}
\paragraph{Off-policy evaluation and historical-data reuse.}
Off-policy evaluation (OPE) provides the statistical foundation for reusing trajectories after a policy changes. Doubly robust estimation combines a learned model with importance weighting to reduce evaluation error \citep{dudik2014doublyrobust}. High-confidence OPE studies how finite samples affect whether a policy comparison can be trusted \citep{thomas2015ref37}, while unequal-support analysis makes explicit when historical data cease to identify the target policy reliably \citep{thomas2017ref36}. More recently, cross-validated OPE has addressed estimator selection under practical finite-data conditions \citep{cief2025crossvalidated}. OPE primarily targets target-policy value; our setting asks a narrower question---whether old evidence still resolves the particular action comparison needed for the current decision.

\paragraph{Pairwise comparison and adaptive evidence acquisition.}
Counterfactual learning shows how logged feedback can support decisions without replaying every alternative \citep{swaminathan2015ref34}. Best-arm and safe-improvement methods similarly allocate evidence toward unresolved comparisons rather than estimating every option equally \citep{simo2019ref30}. Recent agentic reward modeling also treats relative tool behavior as a useful learning target \citep{li2026ref22}. We adopt this comparison-first view, but add a structural condition specific to policy updates: the same global change can matter very differently depending on whether it occurs in states shared by two action branches or in states that distinguish them.

\paragraph{Tool-using agents under change.}
Tool-agent benchmarks increasingly emphasize multistep state consistency and downstream recovery \citep{lu2025toolsandbox}. Tool preferences themselves can remain unstable across prompts and agent configurations \citep{faghih2025toolpreferences}. Continual tool adaptation studies how agents cope when external documentation or tool interfaces evolve \citep{wu2026ref41}, while continual pre-training has been used to improve function calling and adaptation to environmental feedback \citep{zhuang2025hephaestus}. These works establish policy and environment change as recurring features of deployed agents. Our focus is the lifecycle of evidence after such a change: which historical action comparisons remain usable, which can be corrected, and which require fresh execution.

\paragraph{Selective and cost-aware interaction.}
Recent work also treats tool use as a resource-allocation problem. CostBench explicitly evaluates whether agents can avoid unnecessary execution while preserving task performance \citep{liu2026ref25}. Adaptive tool-use methods learn when external tools are worth invoking \citep{li2025ref23}, and SMART targets tool overuse through capability-aware invocation \citep{qian2025smart}. DSC-Gate applies the same selective-interaction principle to credit maintenance. Its decision is whether an update has made historical evidence insufficient enough to justify new target-policy trajectories, connecting OPE-style reuse with pairwise decision sufficiency and execution cost.

\section{Credit Drift and Decision Sufficiency}
\subsection{Pairwise Branch Sensitivity}
Fix a decision context and candidate root actions $a$ and $b$. Let $\mu$ be the old downstream policy and $\pi$ the target policy, with interpolation $\mu_\alpha=\mu+\alpha(\pi-\mu)$. Let $d_{\mu,a}(s,h)$ denote the downstream visitation probability of state $s$ with remaining horizon $h$ after forcing root action $a$. Define the local policy-update effect
\begin{equation}
B_{\mu,\pi}(s,h)=\sum_u [\pi(u\mid s,h)-\mu(u\mid s,h)]Q_\mu(s,h,u).
\end{equation}
The first-order sensitivity of an action pair is
\begin{equation}
G(a,b)=\sum_{s,h}[d_{\mu,a}(s,h)-d_{\mu,b}(s,h)]B_{\mu,\pi}(s,h).
\label{eq:branch-sensitivity}
\end{equation}
This signed inner product captures both \emph{where} the policy changes and \emph{where} the two action-conditioned branches differ. Updates in commonly visited regions can cancel in the pairwise difference, whereas similarly sized updates in branch-differential regions can induce substantial relative credit drift. Exact $G$ is used only in the L1 mechanism analysis; later experiments estimate the needed quantities from finite historical trajectories.

\subsection{First-Order Anchored Credit Transport}
For an old-policy trajectory generated after root action $a$, define the downstream ratio residual
\begin{equation}
z_t=\frac{\pi(A_t\mid S_t,h_t)}{\mu(A_t\mid S_t,h_t)}-1,
\end{equation}
excluding the forced root action from the policy ratio. A first-order expansion of the finite-horizon trajectory likelihood ratio yields an anchored correction. With a state-time baseline $b_t$ fitted on the opposite cross-fitting fold, the trajectory-level quantity is
\begin{equation}
T_i(a)=R_i+\sum_t z_{it}\,[R_i-b_t(S_{it},h_{it})],
\end{equation}
leading to
\begin{equation}
\widehat Q_T(a)=\frac{1}{N}\sum_{i=1}^{N}T_i(a),\qquad
\widehat c_T(a,b)=\widehat Q_T(a)-\widehat Q_T(b).
\label{eq:transport}
\end{equation}
The baseline reduces variance without changing the expectation of the first-order term when fitted independently of the evaluation trajectory. Transport keeps the historical credit as an anchor and estimates only the local policy-induced correction. Its bias grows with the omitted higher-order terms, while its variance depends on the amount and support of the old data. Appendix~\ref{a.2-trajectory-derivation-of-first-order-transport} gives the trajectory derivation and Appendix~\ref{a.3-baselines-and-cross-fitting} details cross-fitting.

\subsection{Ranking Stability}
Let $c_\mu(a,b)=Q_\mu(a)-Q_\mu(b)$ and $\Delta c(a,b)=c_\pi(a,b)-c_\mu(a,b)$. For a non-tied pair, a sufficient condition for preserving the old ranking is
\begin{equation}
\abs{\Delta c(a,b)}<\abs{c_\mu(a,b)}.
\label{eq:ranking-stability}
\end{equation}
This separates three quantities that are often conflated. Branch sensitivity governs the potential relative drift. The old action gap determines whether that drift can alter the choice. Finite-data uncertainty determines whether the updated ranking can be resolved reliably. Large numerical drift may therefore leave the selected action unchanged, while smaller directionally aligned drift can matter near a tight gap.

\begin{figure}[t]
\centering
\includegraphics[width=\linewidth]{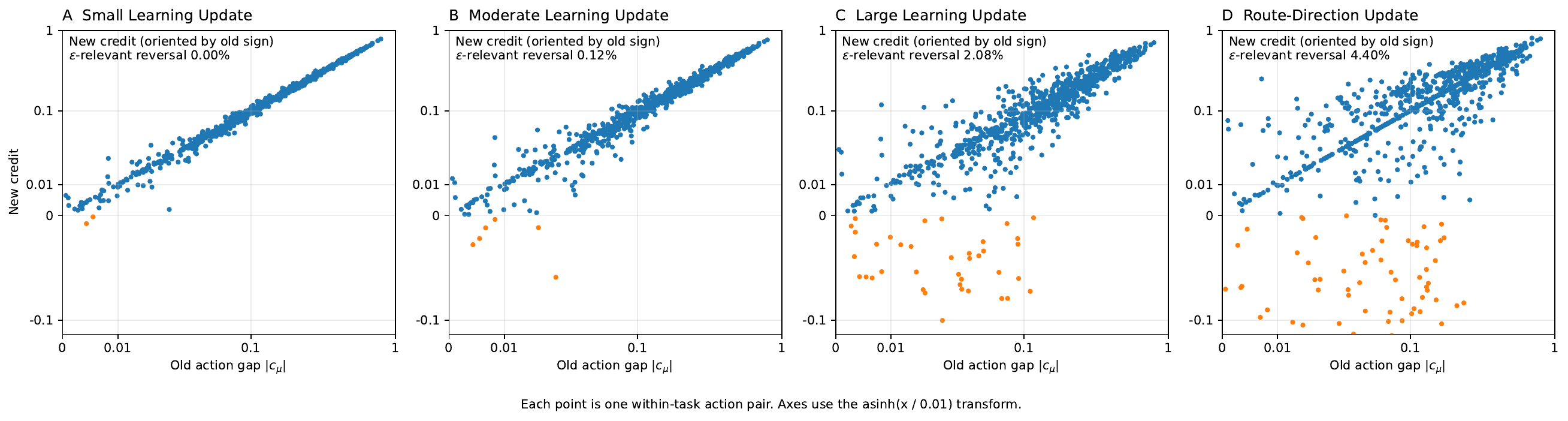}
\caption{\textbf{Credit drift and ranking stability.} Each panel contains all 864 action pairs from the 288 L2--L3 confirmation tasks for a different update condition. Signs are oriented by old credit. Highlighted points are true ranking reversals; the annotated rate additionally requires the absolute new credit difference to exceed the prespecified relevance threshold.}
\label{fig:ranking}
\end{figure}

\subsection{Decision-Sufficient Credit Gate}
Let $\widehat c_R$ be the old action difference from direct reuse and $\widehat\delta$ the first-order correction, with
\begin{equation}
\widehat c_T=\widehat c_R+\widehat\delta.
\end{equation}
We obtain frozen empirical radii $r_R$, $r_\Delta$, and $r_T$ from residual compensation on a development set followed by scale calibration on an independent calibration set. These radii summarize uncertainty in the old gap, the correction, and the transported gap. They are empirical calibration quantities; we do not claim distribution-free coverage or validity under arbitrary stopping.

DSC-Gate resolves a pair in three stages. It \textbf{reuses} historical credit when
\begin{equation}
\abs{\widehat c_R}-r_R>\abs{\widehat\delta}+r_\Delta,
\label{eq:reuse-gate}
\end{equation}
so the conservative old gap exceeds the resolvable drift. If reuse fails, it \textbf{transports} when
\begin{equation}
\abs{\widehat c_T}>r_T,
\label{eq:transport-gate}
\end{equation}
so the transported difference is separated from zero. All remaining pairs enter \textbf{refresh}, where target-policy trajectories are executed until the necessary leader-versus-competitor comparisons are resolved or the 1,536-step cap is reached. With more than two candidate actions, the current leader is repeatedly compared with each competitor, and new execution is allocated only to routes still appearing in unresolved comparisons. The complete gate and allocation protocol is given in Appendix~\ref{d-budgeted-refresh-and-evaluation-algorithm}.

\section{Experimental Design}
We evaluate a sequence of increasingly deployment-oriented questions. The controlled hierarchy L1--L4 uses disjoint task splits where required, fixed seeds, shared old-data budgets, and paired evaluation. The final tool-agent experiment is external validation outside this confirmatory hierarchy. Full environment construction, estimator definitions, calibration, and statistical tests are in Appendices~\ref{b-environment-and-policy-protocol}--\ref{e-statistical-analysis}.

\subsection{L1: Mechanism Validation}
L1 contains 120 development tasks, 120 transfer tasks, and 240 confirmation tasks spanning three semantic families, three workflow sizes, and three horizons. We match global occupancy-weighted KL and apply updates separately to commonly visited regions, branch-differential regions, and random regions. This intervention isolates update location while controlling overall magnitude. Natural finite-sample updates provide a second test, with small steps prespecified for the main analysis and larger steps reserved for boundary analysis. Exact dynamic programming is used only for mechanism tests and evaluation.

\subsection{L2--L3: Finite-Data Estimation and Decisions}
L2--L3 use 480 independently generated tasks split into 96 development, 96 calibration, and 288 confirmation tasks. The confirmation set contains 216 competing-route tasks and 72 serial controls. Each competing task exposes three candidate root routes followed by execution, failure, repair, and submission, with horizons of 6, 9, or 12. A frozen SmolLM2-360M-Instruct model supplies the base behavior policy. Independent sets of 128 trajectories per root action generate the policy updates; evaluation uses nested old-data budgets of 16, 64, and 256 trajectories per action and five evaluation seeds.

We compare direct reuse, first-order credit transport, sequential doubly robust estimation (DR), and self-normalized importance sampling (WIS), together with uniform, adaptive, gap-based, and occupancy-sensitive refresh rules. All methods share the old data and the same latent stream of target-policy trajectories; a method may access only trajectories it has paid for and completed. L2 evaluates pairwise credit MSE. L3 evaluates normalized regret AUC over new-execution budgets from 0 to 1,536 steps. Primary comparisons use within-task pairing, stratified bootstrap resampling, and simultaneous confidence intervals.

\subsection{L4 and Real-Tool Validation}
L4 contains 360 new tasks with no overlap with L1--L3, including 270 competing-route tasks and 90 serial controls. Gate rules, the primary population, and the statistical protocol were frozen before L4 results were read. Each competing task includes ordinary small or moderate learning updates and branch-selective updates constructed from old-policy action-conditioned visitation differences while matching global occupancy-weighted KL. DSC-Gate, Gap-Gate, WIS-Gate, and DR-Gate share the same old data, target policies, calibration quantities, latent new-trajectory stream, and 1,536-step budget. Primary endpoints are terminal regret when the gate stops and actual new tool steps consumed; DSC-Gate is first tested for regret noninferiority to Gap-Gate with margin 0.0005 and then for lower execution cost.

For external validation, we use the retail environment of $\tau^3$-bench with Qwen3-4B-Instruct-2507 as the behavior policy. The target policy is obtained by LoRA-updating the same base model on non-overlapping training trajectories, while the user simulator, tools, system rules, and decoding settings remain fixed. For each test task, three valid root actions are fixed before downstream outcomes are observed, and each receives 64 behavior-policy trajectories. Because exact state visitation is unavailable, we use a trajectory-level first-order branch-sensitivity estimate and compare it with global policy KL. This experiment tests transfer of the mechanism to a real model-weight update; it is not part of the L1--L4 confirmatory hierarchy.

\section{Results}
\subsection{Branch Sensitivity Explains Credit Drift}
After matching global KL, L1 updates concentrated in branch-differential regions produce substantially more pairwise drift than globally similar updates in common regions. Across tasks, the rank-correlation advantage of pairwise sensitivity $\abs{G}$ over global KL for explaining absolute credit drift is 0.839 on average, with a 95\% interval of $[0.830,0.847]$. The first-order correction for local natural updates also passes the prespecified criterion that its error be less than half the error of direct reuse. Similar global update magnitudes therefore need not have similar consequences for a given action comparison.

\begin{figure}[t]
\centering
\includegraphics[width=0.95\linewidth]{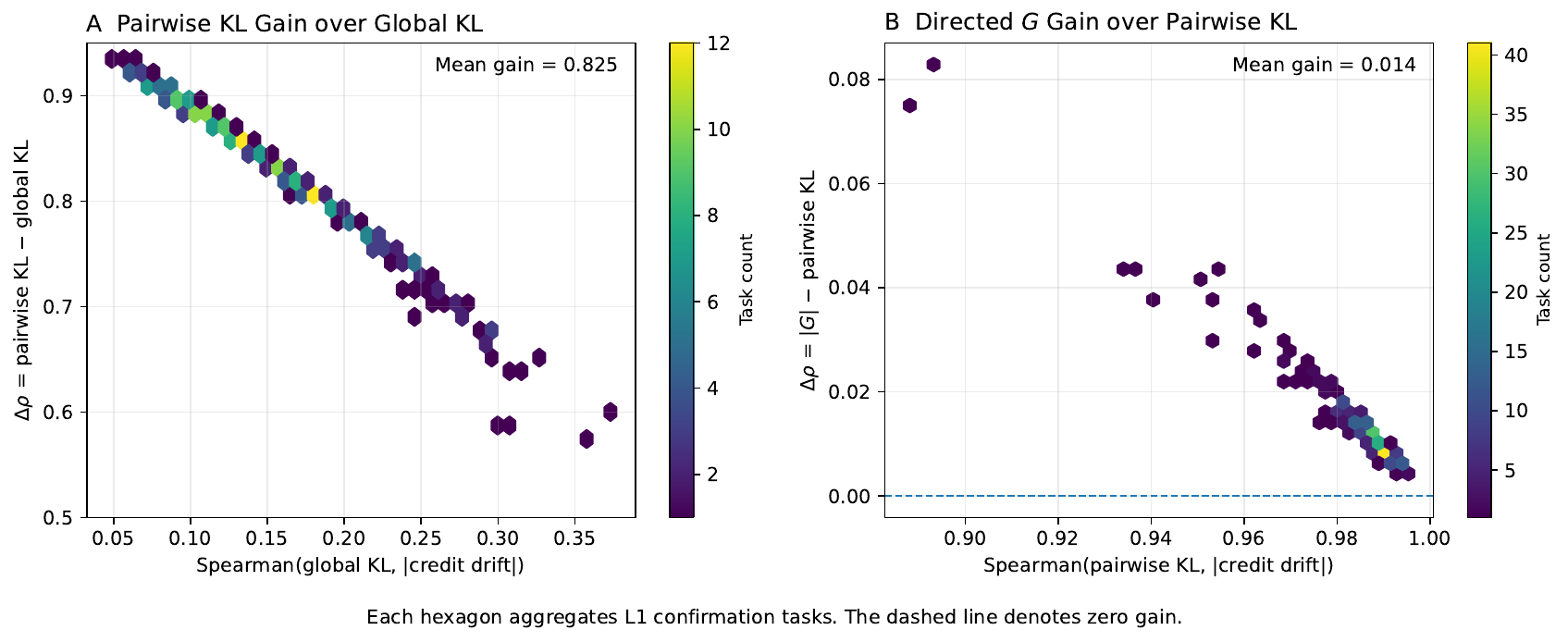}
\caption{\textbf{Branch sensitivity and credit drift in L1.} Panel A shows the gain in Spearman correlation from replacing global KL with pairwise KL. Panel B shows the additional gain from directed branch sensitivity over pairwise KL. Each hexagon aggregates one or more confirmation tasks; the dashed line marks zero gain.}
\label{fig:sensitivity}
\end{figure}

\subsection{Transport Improves Credit Estimates, Selectively Improving Decisions}
With competing routes, small or moderate learning updates, and 256 old trajectories per action, first-order transport yields pairwise credit MSE 0.001225 versus 0.001464 for direct reuse, a 16.3\% reduction. DR reaches 0.001203 in the same slice. The transport advantage grows with old-data volume because the correction becomes less dominated by sampling noise. Under larger updates, the omitted higher-order remainder erodes this first-order advantage.

The decision results reveal a different boundary. In the prespecified L3 analysis, transport with adaptive refresh attains normalized regret AUC 0.006879, an 18.7\% reduction relative to adaptive DR. The simultaneous interval supports this comparison, while the intervals against gap-based and occupancy-sensitive refresh cross zero. True relevance-filtered pairwise reversals are only 0\% and 0.116\% for small and moderate learning updates. Lower credit MSE therefore often leaves the selected action unchanged because the old ranking was already decision-sufficient. Under the single-route direction update, where change is concentrated on branch differences, transport and gap-based refresh obtain regret AUC 0.007897 and 0.011174, respectively, a descriptive 29.3\% reduction.

\begin{table}[t]
\centering
\caption{\textbf{Primary L3 decision comparisons.} Differences are transport-adaptive minus the baseline; lower is better. Intervals are simultaneous for the three prespecified comparisons.}
\label{tab:l3}
\small
\begin{tabular}{lrrl}
\toprule
Baseline & Baseline AUC & Difference & Simultaneous interval \\
\midrule
Adaptive DR & 0.008462 & -0.001583 & [-0.002651, -0.000544] \\
Occupancy-sensitive & 0.006525 & +0.000354 & [-0.000147, +0.000891] \\
Gap-based & 0.006494 & +0.000385 & [-0.000124, +0.000950] \\
\bottomrule
\end{tabular}
\end{table}

\begin{figure}[t]
\centering
\includegraphics[width=\linewidth]{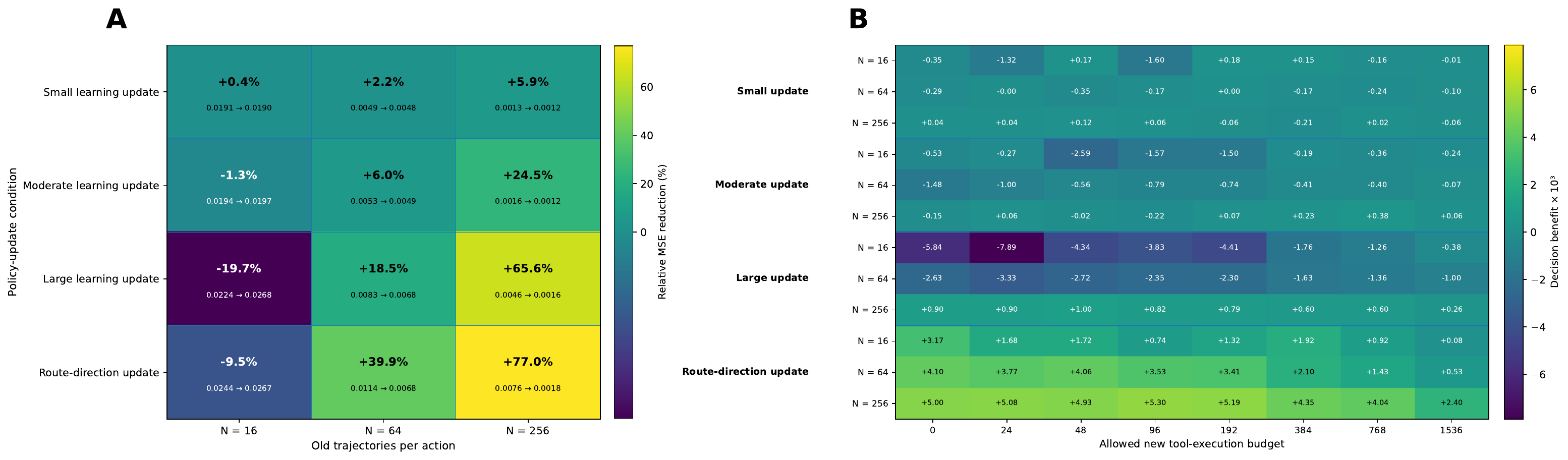}
\caption{\textbf{From credit correction to decision benefit.} Panel A reports the relative reduction in pairwise credit MSE from transport over direct reuse across update conditions and old-data budgets. Panel B reports the corresponding decision benefit across old-data and new-execution budgets. Numerical improvement is most useful when the update can change the action ranking.}
\label{fig:transport}
\end{figure}

\subsection{DSC-Gate Reduces New Execution on the Independent L4 Test}
In the frozen L4 population, mean terminal regret is 0.00673 for DSC-Gate and 0.00669 for Gap-Gate, a difference of $+0.00004$ with paired 95\% interval $[-0.00010,+0.00018]$. The interval's upper bound lies below the prespecified noninferiority margin of 0.0005. At the same time, DSC-Gate uses 286 new tool steps on average, compared with 472 for Gap-Gate, 521 for WIS-Gate, and 558 for DR-Gate, corresponding to reductions of 39.4\%, 45.1\%, and 48.7\%. All three simultaneous cost-ratio intervals lie below 1.

The gate path changes with branch alignment. Under ordinary learning updates, 55.4\% of instances resolve by reuse, 22.8\% by transport, and 21.8\% by refresh. Under branch-selective updates, the proportions shift to 31.8\%, 12.3\%, and 55.9\%. The gate therefore spends new execution where the policy change is concentrated in downstream regions that differentiate the candidate actions.

\begin{table}[t]
\centering
\caption{\textbf{Gate results on the independent L4 test set.}}
\label{tab:l4}
\small
\begin{tabular}{lrrrr}
\toprule
Method & Mean regret & $P(R>0.02)$ & New steps & Any refresh \\
\midrule
DSC-Gate & 0.00673 & 2.9\% & 286 & 38.9\% \\
Gap-Gate & 0.00669 & 3.0\% & 472 & 63.5\% \\
WIS-Gate & 0.00734 & 3.8\% & 521 & 69.4\% \\
DR-Gate & 0.00812 & 4.6\% & 558 & 72.6\% \\
Reuse only & 0.01284 & 8.1\% & 0 & 0\% \\
Transport only & 0.00803 & 4.7\% & 0 & 0\% \\
\bottomrule
\end{tabular}
\end{table}

\subsection{A Real Model-Weight Update Shows the Same Conditional Pattern}
The Qwen3-4B LoRA update raises held-out task success from 36.8\% to 44.2\%, with mean policy KL 0.087 under behavior-policy visitation. Across tasks, trajectory-level branch sensitivity has Spearman correlation 0.58 with reference absolute credit drift, compared with 0.21 for global KL; the difference is $+0.37$ with 95\% bootstrap interval $[0.25,0.48]$. With 64 old trajectories per action, pairwise MSE is 0.0136 for direct reuse and 0.0111 for transport, an 18.4\% reduction; DR and WIS obtain 0.0109 and 0.0124.

DSC-Gate reaches mean terminal regret 0.0291 versus 0.0288 for Gap-Gate while reducing mean new tool steps from 207 to 143, a 30.9\% reduction. Refresh also increases monotonically with branch sensitivity: the lowest sensitivity quartile uses 72 new steps on average and refreshes 15.1\% of instances, whereas the highest quartile uses 224 steps and refreshes 54.7\%. These data provide external-validity evidence for the same mechanism under an actual parameter update and stateful tool use; they remain separate from the L1--L4 confirmatory claims.

\section{Discussion}
The experiments support organizing credit management around the evidence needed for the current action comparison. Pairwise branch sensitivity explains why global policy distance can be a poor proxy for credit expiry: an update matters when it overlaps downstream states that the candidate root actions visit differently. First-order transport then offers a low-cost correction when the change is sufficiently local and the historical data can estimate that correction. The gate adds the remaining ingredient, the action gap, so numerical drift triggers new execution only when it can plausibly change the decision. This distinction also explains why improved credit MSE does not always yield improved regret.

The current scope sets clear boundaries on the claim. DSC-Gate uses empirically calibrated radii; the experiments do not establish distribution-free finite-sample coverage or guarantees under arbitrary adaptive stopping. The reported savings concern target-policy interaction and do not include the historical-data collection, model update, and calibration costs of the full training pipeline. The controlled hierarchy spans several semantic families, workflow scales, horizons, update magnitudes, and branch alignments, while the external validation uses one model configuration and one tool environment after a parameter update. Future work should test repeated model updates, broader tool domains, and longer credit lifecycles, including how historical credit accumulates, transfers, and expires across a sequence of policy changes.

\section{Conclusion}
We study when historical action credit remains sufficient after a policy update. Credit drift depends on how the update aligns with downstream branches, while the need for refresh further depends on the existing action gap and finite-data uncertainty. Pairwise branch sensitivity, anchored credit transport, and DSC-Gate operationalize these factors. On an independent test set, DSC-Gate substantially reduces new tool execution while preserving nearly identical regret, and a real tool-agent parameter update shows the same pattern. The results indicate that historical credit often remains useful after a policy update and can sometimes be corrected from existing trajectories before new execution is required. More broadly, continually updated agents need not rebuild all decision estimates after every policy change. Selective reuse, correction, and refresh can reduce repeated interaction while keeping decisions up to date.

\label{maintextend}

\subsection*{AI Use Statement}
ChatGPT and Codex were used to assist with literature search, language polishing, code editing, translation, and related writing tasks. All AI-assisted content, including factual statements, analyses, and conclusions, was independently reviewed and verified by the authors. The authors take full responsibility for the accuracy and integrity of the manuscript.

\subsection*{Reproducibility Statement}
Appendices A--G document the notation, first-order derivation, environment construction, fixed model and update protocols, estimators, calibration procedure, gate implementation, statistical tests, complete supplementary results, cost accounting, and reproducibility artifacts used in the study. The development, calibration, confirmation, and independent L4 test roles are explicitly separated, and the frozen L4 protocol is described in Appendix~\ref{f.5-independent-l4-gate-test} and Appendix~\ref{g-cost-accounting-and-reproducibility-materials}.

\appendix
\section{Definitions and
Derivations}\label{a-definitions-and-derivations}

\subsection{Notation and Information
Boundaries}\label{a.1-notation-and-information-boundaries}

The context is fixed within each action comparison and omitted from the
notation in the main text. Actions a and b are candidate root actions, $\mu$
is the downstream policy that generated the old data, $\pi$ is the
downstream policy to be deployed, and H is the total horizon including
the root intervention. The terminal return is 1 for a successful
submission and 0 otherwise. $Q_\pi$(a) takes expectation over environment
randomness and subsequent actions under $\pi$ after the root action, and
$c_\pi$(a,b) = $Q_\pi$(a) - $Q_\pi$(b). L1 may contain multiple contexts and action
pairs. Each L2-L3 task fixes three root routes and therefore contains
three unordered action pairs. The task, rather than the trajectory or
action pair, is the basic unit of statistical resampling.

The L2-L3 estimators receive only old trajectories and the known
behavior- and target-policy tables. Dynamic-programming ground truth is
used exclusively for evaluation and error decomposition. Prompts to the
base language model describe stage-level success probabilities, but the
trajectory-estimation and allocation interfaces do not receive simulator
transition tables. The inputs to the behavior policy, the inputs to the
estimator, and the ground truth available to the evaluator must remain
distinct.

\subsection{Trajectory Derivation of First-Order
Transport}\label{a.2-trajectory-derivation-of-first-order-transport}

Assume that the target policy is absolutely continuous on the support of
the old policy and that the environment transition mechanism is
unchanged. Conditional on root intervention a, the environmental
probability factors cancel in the trajectory likelihood ratio. Along the
interpolated policy $\mu_\alpha$, the trajectory ratio is the product of
downstream policy ratios

\[\frac{P_{\text{$\mu_\alpha$}}}{P_{\mu}} = \prod_{t}^{}\left( 1 + \alpha z_{t} \right)\]

\[{Q_{\pi}}_{}(a) = Q_{\mu}(a) + E_{\mu}\lbrack R\sum_{t}^{}z_{t}\rbrack + r_{a}\]

Because the horizon is finite, the finite sum can be differentiated
directly. At $\alpha$ = 0, only terms containing a single z\_t remain, yielding
the directional derivative used in the main text. Setting $\alpha$ = 1 shows
that the higher-order remainder contains all products of two or more
z\_t terms. The remainder depends on both update magnitude and the joint
occurrence of updates along the same trajectory; the first-order term
cannot remove it.

Let g(a) = $\mathbb{E}_\mu${[}R sum\_t z\_t \textbar{} a{]}, so G(a,b) = g(a) - g(b).
Conditioning each directional-derivative term on state and remaining
horizon gives the visitation-difference inner product in the main text.
A commonly visited region cancels only when it has the same weighted
effect on both root branches; common visitation alone does not make the
region invariant.

\subsection{Baselines and
Cross-Fitting}\label{a.3-baselines-and-cross-fitting}

For a baseline b(S\_t,h\_t) fitted independently of the current
evaluation trajectory, the action-conditional expectation satisfies
sum\_u $\mu$(u\textbar s,h){[}$\pi$(u\textbar s,h)/$\mu$(u\textbar s,h)-1{]}b(s,h) =
0 conditional on the state, horizon, and training fold. The baseline
therefore leaves the expectation of the first-order correction
unchanged. In our implementation, b is the old-policy-weighted
state-time action-return estimate fitted on all visited downstream nodes
in the opposite odd-even fold.

Cross-fitting prevents a trajectory from fitting its own baseline,
although the two evaluation folds still share training information, so
the naive variance of the pooled sample remains approximate. We address
practical uncertainty with residual-variance compensation on the
development set and scale factors from the calibration set. This
construction does not imply strict finite-sample coverage.

\subsection{Ranking Stability and
Regret}\label{a.4-ranking-stability-and-regret}

When $c_\mu$ is nonzero, the true ranking is preserved if and only if $c_\mu$$c_\pi$
\textgreater{} 0; equality indicates a tie under the new policy. The
triangle inequality gives \textbar $\Delta c$\textbar{} \textless{}
\textbar $c_\mu$\textbar{} as a sufficient condition. Equivalently, ranking
stability can be written as $\Delta c$/$c_\mu$ \textgreater{} -1 when $c_\mu$ is nonzero.
This ratio is unstable near zero old credit, so Figure 1 displays old
and new credit directly and does not exclude near-ties through the
ratio.

With two actions and no tie, the regret of an incorrect ranking is
\textbar $c_\pi$\textbar. With multiple actions, reversing two suboptimal
actions may leave the optimal choice unchanged. We therefore use true
three-action decision regret rather than pairwise reversal rate as the
decision endpoint. If every estimated action value has absolute error at
most e, the regret of the estimated best action is at most 2e. This is a
general stability fact, not a new certification guarantee introduced
here.

\section{Environment and Policy
Protocol}\label{b-environment-and-policy-protocol}

\subsection{L1 Mechanism
Experiments}\label{b.1-l1-mechanism-experiments}

The confirmatory hypotheses, thresholds, and configuration for L1 are
fixed in protocol.json. The model is HuggingFaceTB/SmolLM2-360M-Instruct
at revision a10cc1512eabd3dde888204e902eca88bddb4951. The development,
transfer, and confirmation splits contain 120, 120, and 240 tasks.
Semantic families are build, data, and publish; workflow sizes are 4, 6,
and 8; horizons are 6, 9, and 12; and each task contains 12 contexts.

The controlled experiment compares differential, common, and random
update supports while matching update magnitude at the same target
global occupancy-weighted KL. The four KL target fractions are 0.15,
0.35, 0.60, and 0.85, the maximum interpolation coefficient is 0.65, and
the support fraction is 0.2. Natural updates use step sizes 0.5, 1, 2,
4, and 8. The first three form the main analysis and the last two probe
the boundary. Update seeds are 11, 33, and 55, with update\_rollouts =
64 and probe\_rollouts = 128. Sign accuracy is computed on eligible
action pairs using the protocol threshold sign\_gap = 0.05 and must not
be conflated with the $\epsilon$ = 0.02 reversal statistic used in L2-L3.

The L1 environment is an n-node directed acyclic workflow. A bit mask
records whether each artifact is valid. A node action is available only
when every parent is valid. Success validates that node and invalidates
all descendants; an unmet precondition or stochastic failure leaves the
state unchanged. Submission terminates immediately and returns 1 only
when every node is valid. A state query consumes one step. Node success
probabilities are sampled independently from \{0.6, 0.8, 1.0\}.
Development tasks are random nonchain DAGs, transfer tasks are chains,
and the confirmation set is split evenly between chains and nonchains.
DAGs are generated in a random topological order with edge probability
0.5, excluding exact chains. Combinations of task family, dependency
structure, and success probabilities do not repeat. Operations
correspond to generating modules, transforming tables, or producing
pages, all governed by the same finite-workflow state structure.

Contexts are drawn from reachable states, four per horizon,
preferentially requiring the shortest possible success path to lie in
{[}2,H{]}. If necessary, remaining contexts are drawn from states from
which success is still possible, and contexts may repeat when fewer than
four candidates are available. Selection depends on feasibility and does
not filter by value gap or method performance. Root actions include
every node action, submission, and state query. The base model reads the
state, action descriptions, and remaining steps, then produces a policy
from constrained action-label probabilities.

The support set is constructed from old-policy occupancy without
querying action values. For each context and unordered action pair, D is
the mean absolute occupancy difference and C is the mean smaller
occupancy. Candidate nodes are state-horizon pairs with global occupancy
above 10$^{-12}$. Support size is ceil(0.2 times the number of candidates),
capped at one third of the candidates. Differential support takes the
largest D values; common support takes the same number from the
remaining nodes ranked by C/(D+10\^{}-9); random support samples without
replacement. On the chosen support, the old policy is mixed with the
tie-uniform greedy policy under exact old Q values. Global KL is KL(new
\textbar\textbar{} old), normalized and weighted by mean downstream
occupancy under the old root-action distribution. Each of the four
targets is a specified fraction of the smallest KL attainable at $\alpha$ =
0.65 across the three support types, and $\alpha$ is matched by 60 steps of
binary search. Pairwise KL uses absolute occupancy difference to weight
local KL and is not normalized in the same way.

Natural updates sample a return table Binomial(64,$Q_\mu$)/64 for every
action at every candidate node, then form a mirror-descent update
proportional to $\mu$(u\textbar s,h) exp{[}$\eta$ Â(s,h,u){]}. An independent
probe table Binomial(128,$Q_\mu$)/128 supplies local value changes for
sampled G. Occupancy and the old-credit anchor remain exact. Sampled G
is therefore a model-assisted diagnostic with finite return noise,
rather than a pure trajectory-based off-policy estimator or evidence of
realized execution savings. L2-L3 test the implementable procedure using
transition-by-transition old trajectories and paid new data.

\subsection{L2-L3 Workflows}\label{b.2-l2-l3-workflows}

The task-generation seed is 270917. Within each split, the three task
families and horizons are assigned in a crossed design, and one quarter
of tasks are serial controls. A competing-route task independently
samples three route lengths, each containing two to four stages. Fast,
careful, and repair success probabilities at each stage are sampled
uniformly from {[}0.35,0.94{]}, {[}0.55,0.96{]}, and {[}0.55,0.98{]}.
Serial controls have route lengths 2, 3, and 4, share one set of
probabilities across all stages, and draw the three probabilities from
{[}0.48,0.90{]}, {[}0.60,0.95{]}, and {[}0.65,0.98{]}. Route indices are
randomly permuted, and tasks are not filtered by exact value.

In a normal stage, the agent can execute quickly or carefully. Success
advances one stage; fast failure causes damage, whereas careful failure
leaves the state unchanged. In a damaged state, repair is the only valid
action. Successful repair restores the normal state, and failed repair
leaves the state damaged. Submission becomes valid only after all stages
are complete. Every tool consumes one step. A successful submission
returns 1, and horizon exhaustion returns 0. Choosing the root route
also consumes one step, so the shortest complete trajectory costs the
number of route stages plus two. Shared dynamics in the serial controls
do not imply that rankings remain exactly invariant after an arbitrary
finite-sample learning update.

\subsection{Base Policy and Independent
Updates}\label{b.3-base-policy-and-independent-updates}

The base-model revision matches L1. Given a static state snapshot and a
valid-tool mask, the model outputs action propensities that are mixed
with 15\% uniform exploration over valid actions. It does not observe
the remaining horizon. Downstream policy tables are indexed by remaining
horizon and can therefore represent time-dependent tabular learning
updates.

Each root route independently generates 128 trajectories under $\mu$ for
policy updating; these trajectories are separate from the
credit-estimation data. Terminal returns are aggregated by state,
remaining horizon, and action. Action value is estimated as (successes +
0.5)/(visits + 1), with unseen actions reverting to 0.5. The estimated
advantage is multiplied by intensity $\eta$, used to multiplicatively
reweight $\mu$ through a softmax, and then mixed with 10\% $\mu$. Values $\eta$ =
0.7, 1.5, and 4 define small, moderate, and large updates.

For the single-route direction condition, the route and sign are fixed
during task generation. Only that route receives log-weight offsets of
+/-0.75 on fast and careful actions, followed by the same 10\% mixture
with the old policy. This condition isolates directional effects. It is
not selected from observed gains and does not replace evidence from
learning updates. The language-model weights remain fixed throughout
these experiments.

\section{Estimators and
Uncertainty}\label{c-estimators-and-uncertainty}

\subsection{Shared Data and Four
Estimators}\label{c.1-shared-data-and-four-estimators}

Evaluation seeds are 11, 22, 33, 44, and 55. For each task, seed, and
route, we store up to 256 old trajectories; budgets of 16, 64, and 256
are nested prefixes. Each trajectory records downstream states, actions,
behavior probabilities, terminal return, and length including the root
action. The two-fold baseline uses visited nodes from all routes without
sampling additional hidden-state returns.

Direct reuse estimates $Q_\mu$(a) by its sample mean and substitutes that
estimate for $Q_\pi$(a). Transport adds the anchored correction from the main
text. Sequential DR begins at the baseline target-policy state value and
sums the cumulative importance weight times immediate return plus
next-state baseline value minus current-action baseline value. The
baseline action value is fitted from $\mu$ data, and the state baseline is
weighted under $\pi$. With exact ratios and independent fitting, telescoping
does not require the baseline to equal $Q_\pi$ exactly.

WIS assigns each trajectory the product W of all downstream ratios and
estimates value as sum(WR)/sum(W). Its finite-sample bias is retained in
evaluation. Effective sample size is (sum W)\^{}2/sum(W\^{}2). WIS
variance is the sample variance of the normalized influence quantity W(R
- estimated mean)/mean(W), divided by N. Other estimators use the sample
variance of their trajectory-level quantities divided by N. Every
estimator includes a variance floor of 1/(4N\^{}2). Raw L2 predictions
are not clipped.

\subsection{Development and
Calibration}\label{c.2-development-and-calibration}

For every update condition and root route, development and calibration
tasks each receive 1,024 target-policy reference trajectories. On the
development set, squared prediction error against the reference value is
computed separately by old-data budget and estimator. The estimated
sampling and reference variances are subtracted, and the remainder is
clipped at zero to form additional residual variance. This calculation
uses only the small and moderate primary updates, after which it is
frozen for all conditions.

For action pair \(\left( \mathbf{a},\mathbf{b} \right)\), define the old
credit, true credit drift, and target credit as

\[\mathbf{c}_{\mathbf{\mu}}\left( \mathbf{a},\mathbf{b} \right),\ \Delta\mathbf{c}\left( \mathbf{a},\mathbf{b} \right) = \mathbf{c}_{\mathbf{\pi}}\left( \mathbf{a},\mathbf{b} \right) - \mathbf{c}_{\mathbf{\mu}}\left( \mathbf{a},\mathbf{b} \right),\ \mathbf{c}_{\mathbf{\pi}}\left( \mathbf{a},\mathbf{b} \right).
\]

The corresponding estimates are direct reuse
\({\widehat{\mathbf{c}}}_{\mathbf{R}}\), first-order correction
\(\widehat{\mathbf{\delta}}\), and transported credit

\[{\widehat{\mathbf{c}}}_{\mathbf{T}} = {\widehat{\mathbf{c}}}_{\mathbf{R}} + \widehat{\mathbf{\delta}}.
\]

The three errors are therefore

\[\mathbf{e}_{\mathbf{R}} = {\widehat{\mathbf{c}}}_{\mathbf{R}} - \mathbf{c}_{\mathbf{\mu}},\ \mathbf{e}_{\Delta} = \widehat{\mathbf{\delta}} - \Delta\mathbf{c},\ \mathbf{e}_{\mathbf{T}} = {\widehat{\mathbf{c}}}_{\mathbf{T}} - \mathbf{c}_{\mathbf{\pi}}.
\]

The development set separately estimates the residual variance in each
error that is not explained by analytic sampling variance, and these
estimates are then frozen for calibration. Here \(\mathbf{e}_{\Delta}\)
contains both finite-sample error and the higher-order remainder of the
first-order approximation.

On the independent calibration set, each error is standardized by its
compensated standard error, and the maximum is taken over the
prespecified action pairs, primary update conditions, and evaluation
seeds within each task. At nominal level 0.95, the following order
statistic is selected from the 96 calibration tasks

\[\left\lceil \left( \mathbf{96} + \mathbf{1} \right) \times \mathbf{0}.\mathbf{95} \right\rceil = \mathbf{93}
\]

yielding the frozen scale factors

\[\mathbf{\kappa}_{\mathbf{R}},\ \mathbf{\kappa}_{\Delta},\ \mathbf{\kappa}_{\mathbf{T}}.
\]

These factors are used for both the empirical acceptance diagnostic in
L2 and the decision radii in DSC-Gate. Because reference quantities
contain sampling noise, cross-fitted estimates are dependent, and
first-order transport includes a higher-order remainder, we interpret
the procedure as empirical calibration and do not claim formal coverage.

\subsection{Decision-Sufficiency
Radii}\label{c.3-decision-sufficiency-radii}

Using the frozen scale factors from Section C.2, define
\(r_{R} = \kappa_{R}s_{R},\ r_{\Delta} = \kappa_{\Delta}s_{\Delta},\ r_{T} = \kappa_{T}s_{T},\)
where \(s_{R},s_{\Delta},s_{T}\) are the compensated standard errors for
\({\widehat{c}}_{R},\widehat{\delta},{\widehat{c}}_{T}\), respectively.
Because \(\widehat{\delta} = {\widehat{c}}_{T} - {\widehat{c}}_{R},\)
computation of \(s_{\Delta}\) retains the covariance induced by the old
trajectories shared between \({\widehat{c}}_{T}\) and
\({\widehat{c}}_{R}\). The three radii correspond to
\(\mid {\widehat{c}}_{R} - c_{\mu} \mid ,\  \mid \widehat{\delta} - \Delta c \mid ,\  \mid {\widehat{c}}_{T} - c_{\pi} \mid\).

When
\(\mid {\widehat{c}}_{R} \mid - r_{R} > \mid \widehat{\delta} \mid + r_{\Delta}\),
the empirical lower bound on the old action gap exceeds the empirical
upper bound on credit drift. If the corresponding empirical error events
hold, this condition implies
\(\mid c_{\mu} \mid > \mid \Delta c \mid ,\) and therefore preserves the
old ranking by Section A.4. If it fails but
\(\mid {\widehat{c}}_{T} \mid > r_{T},\), the transported action
difference is separated from zero and determines the ranking. All
remaining comparisons enter the refresh set.

These radii define a frozen empirical decision rule. They do not provide
distribution-free coverage or coverage under arbitrary stopping.

\section{Budgeted Refresh and Evaluation
Algorithm}\label{d-budgeted-refresh-and-evaluation-algorithm}

\subsection{Old-Data Priors and
Sampling}\label{d.1-old-data-priors-and-sampling}

For route a and estimator e, let m be the raw estimated mean, v the
variance after development-set residual compensation, and N the old
sample size. The pseudo-count is min(N, 0.25/max(v,10$^{-12}$)), and prior
successes equal the pseudo-count times m clipped to {[}0,1{]}. Successes
and counts from newly completed trajectories are added to this prior.
Beta(0.5,0.5) smoothing then gives the posterior mean and approximate
variance.

The uniform rule prioritizes the route with the fewest completed new
trajectories. The adaptive rule uses the current leader as reference,
smooths each estimated gap by max(estimated gap, 0.02), and allocates in
proportion to posterior variance divided by squared gap. For the leader,
the gap to its closest competitor is used. The occupancy-sensitive
baseline further multiplies this score by 1 plus pairwise local KL
divided by mean pairwise local KL, where local KL is estimated from
downstream visitation in the old trajectories. Sampling-score ties are
broken at random. Final decisions are distributed uniformly among
posterior maxima within a tolerance of 10$^{-12}$.

Uniform and adaptive transport share the same prior construction, as do
the two DR variants. Gap-based and occupancy-sensitive refresh use the
reuse prior. The two fresh variants use no old pseudo-data. Reuse-only
selects from the old mean and always spends zero new steps. WIS uses
adaptive allocation as an additional strong baseline.

\subsection{Execution and Information
Isolation}\label{d.2-execution-and-information-isolation}

Algorithm 1 receives the old trajectories, $\mu$ and $\pi$, frozen calibration
quantities, and the allowed budget grid. Cross-fitting first produces
route-level means and variances for each estimator, which initialize the
corresponding priors. In each round, the method selects a route using
only the currently visible posterior and runs the target policy until
the trajectory ends or the budget is exhausted. Transitions from an
incomplete trajectory count against the budget, but its terminal outcome
does not update the posterior. The current decision distribution and
actual expenditure are saved at every budget point before execution
continues. Only completed trajectories update the posterior. An
independent evaluator finally computes regret and expected success.

All methods share a latent stream of route trajectories to reduce
simulation noise in paired comparisons. Route length and future outcomes
are unavailable before a sampling decision. Unpurchased trajectories in
the pool are not learner data. Trajectories are generated through actual
stepwise environment transitions rather than Bernoulli draws from exact
Q.

\subsection{Decision-Sufficiency and Baseline
Gates}\label{d.3-decision-sufficiency-and-baseline-gates}

Using only old trajectories, DSC-Gate first computes the reuse
difference
\({\widehat{\mathbf{c}}}_{\mathbf{R}}\left( \mathbf{a},\mathbf{b} \right)\),
first-order correction
\(\widehat{\mathbf{\delta}}\left( \mathbf{a},\mathbf{b} \right)\),
transported difference
\({\widehat{\mathbf{c}}}_{\mathbf{T}}\left( \mathbf{a},\mathbf{b} \right) = {\widehat{\mathbf{c}}}_{\mathbf{R}}\left( \mathbf{a},\mathbf{b} \right) + \widehat{\mathbf{\delta}}\left( \mathbf{a},\mathbf{b} \right)\),
and empirically calibrated radii
\(\mathbf{r}_{\mathbf{R}},\mathbf{r}_{\Delta},\mathbf{r}_{\mathbf{T}}\).
These old-data quantities remain fixed throughout refresh. For each
competitor to the current leading route, if

\[\mid {\widehat{\mathbf{c}}}_{\mathbf{R}} \mid - \mathbf{r}_{\mathbf{R}} > \mid \widehat{\mathbf{\delta}} \mid + \mathbf{r}_{\mathbf{\Delta}},
\]

the comparison is marked reuse-resolved. If that condition fails but

\[\mid {\widehat{\mathbf{c}}}_{\mathbf{T}} \mid > \mathbf{r}_{\mathbf{T}},
\]

the comparison is marked transport-resolved. Every other comparison
enters the refresh set.

For gate \(\mathbf{e}\), let
\(\mathbf{m}_{\mathbf{t},\mathbf{e}}\left( \mathbf{a} \right)\) and
\(\mathbf{v}_{\mathbf{t},\mathbf{e}}\left( \mathbf{a} \right)\) denote
the posterior mean and approximate variance of route a after refresh
round \(\mathbf{t}\). For an action pair in the refresh set, define

\[{\widehat{\mathbf{c}}}_{\mathbf{t},\mathbf{e}}^{\mathbf{F}}\left( \mathbf{a},\mathbf{b} \right) = \mathbf{m}_{\mathbf{t},\mathbf{e}}\left( \mathbf{a} \right) - \mathbf{m}_{\mathbf{t},\mathbf{e}}\left( \mathbf{b} \right).
\]

Under the route-wise Beta posterior approximation in Section D.1, route
posteriors are treated as conditionally independent, so the approximate
standard error of the action difference is

\[\mathbf{s}_{\mathbf{t},\mathbf{e}}^{\mathbf{F}}\left( \mathbf{a},\mathbf{b} \right) = \sqrt{\mathbf{v}_{\mathbf{t},\mathbf{e}}\left( \mathbf{a} \right) + \mathbf{v}_{\mathbf{t},\mathbf{e}}\left( \mathbf{b} \right)}.
\]

Using the frozen empirical scale factor \(\mathbf{\kappa}_{\mathbf{e}}\)
for the corresponding estimator from Section C.2, define the refresh
radius as

\[\mathbf{r}_{\mathbf{t},\mathbf{e}}^{\mathbf{F}}\left( \mathbf{a},\mathbf{b} \right) = \mathbf{\kappa}_{\mathbf{e}}\mathbf{s}_{\mathbf{t},\mathbf{e}}^{\mathbf{F}}\left( \mathbf{a},\mathbf{b} \right).
\]

When

\[\mid {\widehat{\mathbf{c}}}_{\mathbf{t},\mathbf{e}}^{\mathbf{F}}\left( \mathbf{a},\mathbf{b} \right) \mid > \mathbf{r}_{\mathbf{t},\mathbf{e}}^{\mathbf{F}}\left( \mathbf{a},\mathbf{b} \right)
\]

the comparison is marked refresh-resolved, with direction determined by
the sign of
\({\widehat{\mathbf{c}}}_{\mathbf{t},\mathbf{e}}^{\mathbf{F}}\left( \mathbf{a},\mathbf{b} \right)\).
Like the old-data radii, this criterion is an empirically calibrated
stopping rule and carries no distribution-free finite-sample coverage
guarantee.

Unresolved action pairs form a refresh graph, and only routes incident
to that graph may receive new target-policy trajectories. Route
allocation follows the variance-over-squared-gap rule from Section D.1.
After each new trajectory completes, only that route\textquotesingle s
posterior mean and variance are updated. The current leader is then
recomputed and compared with every competitor using reuse, transport, or
refresh evidence. Execution stops as soon as every necessary comparison
is resolved. At \(\mathbf{B}_{\max} = \mathbf{1536}\), the procedure
stops forcibly and selects from the current posterior.

\subsection{Metrics}\label{d.4-metrics}

For L2 MSE, the three action pairs are first averaged within each task
and then averaged over seeds and the specified conditions. An $\epsilon$-relevant
true reversal requires $c_\mu$$c_\pi$ \textless{} 0 and \textbar $c_\pi$\textbar{}
\textgreater{} 0.02. This statistic measures a change in true value and
is distinct from an incorrect choice caused by estimation noise. The
unseen fraction records the share of state-time-action tuples visited by
evaluation trajectories that do not appear in the opposite training
fold. The ESS ratio is normalized by N.

For each L3 task, update, seed, old-data budget, method, and
new-execution budget, we save five metrics: expected regret, probability
of selecting an action with regret above 0.02, expected success, new
tool steps consumed, and new trajectories completed. The curve is
integrated by the trapezoidal rule over log(1+B) and divided by the
total horizontal width log(1537), yielding normalized AUC. L4 does not
integrate the full budget curve. Instead, it records expected regret,
P(regret \textgreater{} 0.02), actual new tool steps, completed
trajectories, any refresh, budget-cap incidence, and the final
proportions of comparisons resolved by reuse, transport, and refresh
when each gate stops.

\section{Statistical Analysis}\label{e-statistical-analysis}

The three prespecified L1 statistics are task-level
log(MSE\_differential/MSE\_common) - log(1.25);
Spearman(\textbar G\textbar,\textbar drift\textbar) - Spearman(global
KL,\textbar drift\textbar) - 0.1; and MSE\_first-order - 0.5 MSE\_reuse.
Each uses 10,000 task-level bootstrap resamples of the 240 confirmation
tasks. The respective passing criteria are a lower interval bound above
zero, a lower bound above zero, and an upper bound below zero. The mean
task-level log ratio differs from the ratio of aggregated MSEs, so the
L1-A value 6.7068 must not be described directly as a multiplicative
drift factor.

The primary L3 population comprises 216 competing-route confirmation
tasks, equally weighting small and moderate learning updates, three
old-data budgets, and five seeds. Method differences are formed within
the same task and seed. Tasks are then resampled within task-family and
control-type strata, while seeds are resampled globally with shared
indices to preserve pairing. We perform 10,000 bootstrap resamples. Each
of the three primary comparisons uses two-sided $\alpha$ = 0.05/3,
corresponding to quantiles 0.05/6 and 1 - 0.05/6, to form Bonferroni
simultaneous intervals. The complete confirmation set contains 288
tasks; action pairs and trajectories are not treated as independent
confirmation samples.

The original directional conclusion for L3 remains unchanged. Among the
simultaneous intervals comparing transport with DR, occupancy-sensitive
refresh, and gap-based refresh, only the DR comparison lies entirely
below zero. The update-slice heat map, the 29.3\% gain under direction
updates, and the 16.3\% numerical gain at the largest old-data budget
are descriptive point estimates.

L4 uses an independent task set with no overlap with L1-L3 and defines a
new confirmatory question. The prespecified primary population equally
weights ordinary and branch-selective small or moderate learning updates
on 270 competing-route tasks, then equally weights the three old-data
budgets and five evaluation seeds. The first step tests whether terminal
regret under DSC-Gate is noninferior to Gap-Gate at a fixed margin of
0.0005. Only after passing that test do we test whether DSC-Gate uses
fewer new tool steps than Gap-Gate, WIS-Gate, and DR-Gate
simultaneously. The original task is the bootstrap unit. Whenever a task
is resampled, its ordinary and branch-selective updates remain paired.
Tasks are resampled within task-family strata, and evaluation seeds use
shared global indices. We perform 10,000 paired bootstrap resamples.
After the regret noninferiority test passes, the three cost ratios are
compared with 1 using Bonferroni simultaneous intervals.

\section{Complete Results and Supplementary
Analyses}\label{f-complete-results-and-supplementary-analyses}

\subsection{L1 Confirmation Tests and Large
Updates}\label{f.1-l1-confirmation-tests-and-large-updates}

Table A1. Prespecified L1 confirmation tests. Statistics are defined in
Appendix E; n = 240 for each test.

\begin{longtable}[]{@{}
  >{\raggedright\arraybackslash}p{(\columnwidth - 6\tabcolsep) * \real{0.1177}}
  >{\raggedright\arraybackslash}p{(\columnwidth - 6\tabcolsep) * \real{0.2353}}
  >{\raggedright\arraybackslash}p{(\columnwidth - 6\tabcolsep) * \real{0.4705}}
  >{\raggedright\arraybackslash}p{(\columnwidth - 6\tabcolsep) * \real{0.1765}}@{}}
\toprule\noalign{}
\begin{minipage}[b]{\linewidth}\raggedright
Test
\end{minipage} & \begin{minipage}[b]{\linewidth}\raggedright
Mean statistic
\end{minipage} & \begin{minipage}[b]{\linewidth}\raggedright
95\% interval
\end{minipage} & \begin{minipage}[b]{\linewidth}\raggedright
Result
\end{minipage} \\
\midrule\noalign{}
\endhead
\bottomrule\noalign{}
\endlastfoot
L1-A & 6.706786 & {[}5.872131, 7.521799{]} & Passed \\
L1-B & 0.7387006 & {[}0.7298759, 0.7472724{]} & Passed \\
L1-C & -0.001264017 & {[}-0.001395845, -0.001137836{]} & Passed \\
\end{longtable}

\begin{figure}[t]
\centering
\includegraphics[width=\linewidth]{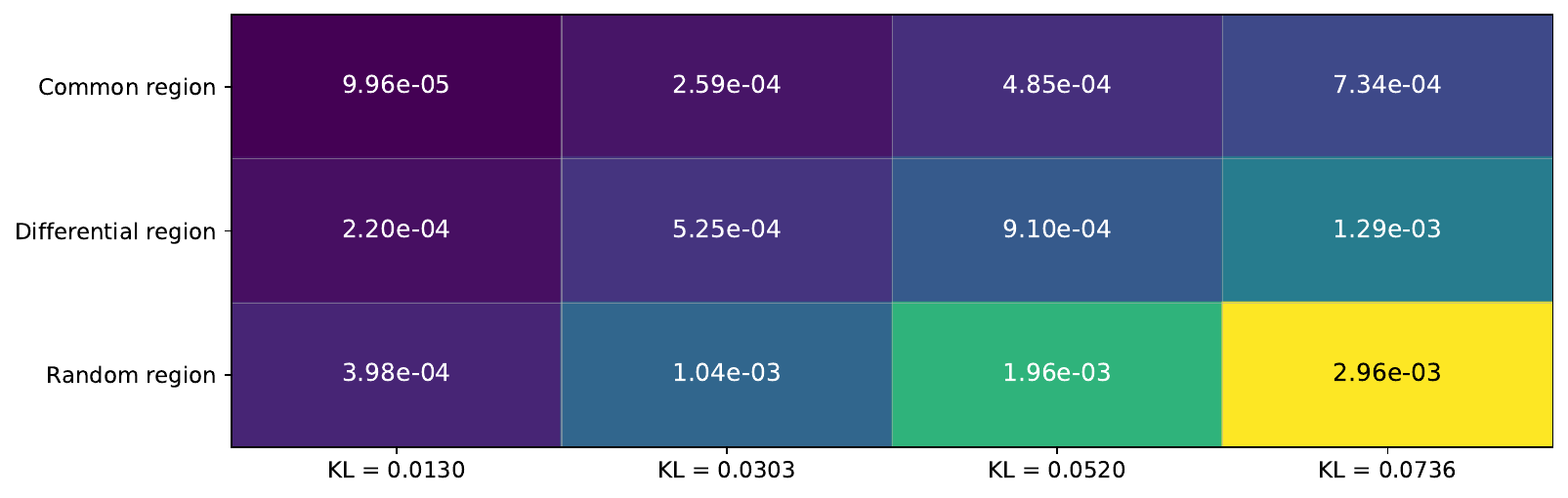}
\end{figure}

Figure A1. Matched-KL results for controlled updates. Columns are update
levels at matched global occupancy-weighted KL; cells contain mean
pairwise drift MSE across confirmation tasks. Common, differential, and
random regions are all retained. This aggregate is shown for
interpretation; the formal L1-A statistic remains the task-level log
ratio.

Table A2. L1 natural-update boundary by step size. Each row contains 240
tasks times 3 update seeds.

\begin{longtable}[]{@{}
  >{\raggedright\arraybackslash}p{(\columnwidth - 8\tabcolsep) * \real{0.0882}}
  >{\raggedright\arraybackslash}p{(\columnwidth - 8\tabcolsep) * \real{0.2235}}
  >{\raggedright\arraybackslash}p{(\columnwidth - 8\tabcolsep) * \real{0.2353}}
  >{\raggedright\arraybackslash}p{(\columnwidth - 8\tabcolsep) * \real{0.2471}}
  >{\raggedright\arraybackslash}p{(\columnwidth - 8\tabcolsep) * \real{0.2059}}@{}}
\toprule\noalign{}
\begin{minipage}[b]{\linewidth}\raggedright
Step size
\end{minipage} & \begin{minipage}[b]{\linewidth}\raggedright
Reuse MSE
\end{minipage} & \begin{minipage}[b]{\linewidth}\raggedright
Exact first-order MSE
\end{minipage} & \begin{minipage}[b]{\linewidth}\raggedright
Sampled first-order MSE
\end{minipage} & \begin{minipage}[b]{\linewidth}\raggedright
Eligible-pair reversal rate
\end{minipage} \\
\midrule\noalign{}
\endhead
\bottomrule\noalign{}
\endlastfoot
0.5 & 3.750e-04 & 1.443e-06 & 1.518e-06 & 0.0000\% \\
1 & 1.688e-03 & 3.261e-05 & 3.288e-05 & 0.0000\% \\
2 & 7.336e-03 & 8.734e-04 & 8.739e-04 & 0.0000\% \\
4 & 2.180e-02 & 1.421e-02 & 1.420e-02 & 0.0040\% \\
8 & 3.912e-02 & 4.105e-02 & 4.106e-02 & 0.0386\% \\
\end{longtable}

The table reports mean MSE at each natural-update step size, not the
mean task-level relative improvement. At step size 8, task-averaged
first-order MSE exceeds reuse MSE, showing that the local approximation
cannot be extrapolated directly to large updates. This result is
compatible with a positive task-level relative improvement because the
two summaries weight tasks differently.

\subsection{All L2 Estimators and Noise
Diagnostics}\label{f.2-all-l2-estimators-and-noise-diagnostics}

Table A3. Pairwise MSE on competing routes for all update conditions and
old-data budgets. Predictions are raw and unclipped.

\begin{longtable}[]{@{}
  >{\raggedright\arraybackslash}p{(\columnwidth - 10\tabcolsep) * \real{0.1530}}
  >{\raggedright\arraybackslash}p{(\columnwidth - 10\tabcolsep) * \real{0.0941}}
  >{\raggedright\arraybackslash}p{(\columnwidth - 10\tabcolsep) * \real{0.1882}}
  >{\raggedright\arraybackslash}p{(\columnwidth - 10\tabcolsep) * \real{0.1882}}
  >{\raggedright\arraybackslash}p{(\columnwidth - 10\tabcolsep) * \real{0.1882}}
  >{\raggedright\arraybackslash}p{(\columnwidth - 10\tabcolsep) * \real{0.1882}}@{}}
\toprule\noalign{}
\begin{minipage}[b]{\linewidth}\raggedright
Update
\end{minipage} & \begin{minipage}[b]{\linewidth}\raggedright
Old N
\end{minipage} & \begin{minipage}[b]{\linewidth}\raggedright
Reuse
\end{minipage} & \begin{minipage}[b]{\linewidth}\raggedright
Transport
\end{minipage} & \begin{minipage}[b]{\linewidth}\raggedright
DR
\end{minipage} & \begin{minipage}[b]{\linewidth}\raggedright
WIS
\end{minipage} \\
\midrule\noalign{}
\endhead
\bottomrule\noalign{}
\endlastfoot
Small learning update & 16 & 0.019084 & 0.019003 & 0.025226 &
0.018989 \\
Small learning update & 64 & 0.004864 & 0.004758 & 0.005483 &
0.004767 \\
Small learning update & 256 & 0.001289 & 0.001213 & 0.001199 &
0.001216 \\
Moderate learning update & 16 & 0.019444 & 0.019702 & 0.026032 &
0.019536 \\
Moderate learning update & 64 & 0.005250 & 0.004933 & 0.005670 &
0.004952 \\
Moderate learning update & 256 & 0.001639 & 0.001238 & 0.001206 &
0.001250 \\
Large learning update & 16 & 0.022422 & 0.026844 & 0.035191 &
0.025206 \\
Large learning update & 64 & 0.008317 & 0.006775 & 0.007897 &
0.006962 \\
Large learning update & 256 & 0.004622 & 0.001591 & 0.001514 &
0.001705 \\
Route-direction update & 16 & 0.024356 & 0.026677 & 0.044400 &
0.026311 \\
Route-direction update & 64 & 0.011379 & 0.006842 & 0.010116 &
0.008078 \\
Route-direction update & 256 & 0.007637 & 0.001754 & 0.001866 &
0.002044 \\
\end{longtable}

Table A4. Empirical acceptance and data-support diagnostics for the
primary updates. Small and moderate updates are equally weighted;
post-acceptance error is the aggregated error numerator divided by the
aggregated acceptance rate.

\begin{longtable}[]{@{}
  >{\raggedright\arraybackslash}p{(\columnwidth - 10\tabcolsep) * \real{0.0765}}
  >{\raggedright\arraybackslash}p{(\columnwidth - 10\tabcolsep) * \real{0.1529}}
  >{\raggedright\arraybackslash}p{(\columnwidth - 10\tabcolsep) * \real{0.1882}}
  >{\raggedright\arraybackslash}p{(\columnwidth - 10\tabcolsep) * \real{0.2177}}
  >{\raggedright\arraybackslash}p{(\columnwidth - 10\tabcolsep) * \real{0.1882}}
  >{\raggedright\arraybackslash}p{(\columnwidth - 10\tabcolsep) * \real{0.1765}}@{}}
\toprule\noalign{}
\begin{minipage}[b]{\linewidth}\raggedright
Old N
\end{minipage} & \begin{minipage}[b]{\linewidth}\raggedright
Estimator
\end{minipage} & \begin{minipage}[b]{\linewidth}\raggedright
Acceptance rate
\end{minipage} & \begin{minipage}[b]{\linewidth}\raggedright
Post-acceptance error
\end{minipage} & \begin{minipage}[b]{\linewidth}\raggedright
Unseen fraction
\end{minipage} & \begin{minipage}[b]{\linewidth}\raggedright
ESS/N
\end{minipage} \\
\midrule\noalign{}
\endhead
\bottomrule\noalign{}
\endlastfoot
16 & Reuse & 5.74\% & 0.000\% & 23.36\% & 0.976 \\
16 & Transport & 5.57\% & 0.000\% & 23.36\% & 0.976 \\
16 & DR & 4.31\% & 0.000\% & 23.36\% & 0.976 \\
16 & WIS & 4.32\% & 0.000\% & 23.36\% & 0.976 \\
64 & Reuse & 30.43\% & 0.000\% & 5.16\% & 0.975 \\
64 & Transport & 25.76\% & 0.000\% & 5.16\% & 0.975 \\
64 & DR & 25.08\% & 0.000\% & 5.16\% & 0.975 \\
64 & WIS & 25.11\% & 0.000\% & 5.16\% & 0.975 \\
256 & Reuse & 42.41\% & 0.000\% & 0.53\% & 0.974 \\
256 & Transport & 46.84\% & 0.000\% & 0.53\% & 0.974 \\
256 & DR & 44.55\% & 0.000\% & 0.53\% & 0.974 \\
256 & WIS & 46.53\% & 0.000\% & 0.53\% & 0.974 \\
\end{longtable}

Acceptance and post-acceptance ranking error should be read together, so
a low error obtained by accepting only a few cases is not mistaken for
broad reuse reliability. The unseen fraction and ESS are data-support
diagnostics and do not constitute formal coverage guarantees.

\subsection{All L3 Methods and Update
Conditions}\label{f.3-all-l3-methods-and-update-conditions}

Table A5. Normalized regret AUC for all methods on competing routes.
Old-data budgets and seeds are equally weighted; the primary analysis
averages small and moderate updates.

\begin{longtable}[]{@{}
  >{\raggedright\arraybackslash}p{(\columnwidth - 10\tabcolsep) * \real{0.1882}}
  >{\raggedright\arraybackslash}p{(\columnwidth - 10\tabcolsep) * \real{0.1588}}
  >{\raggedright\arraybackslash}p{(\columnwidth - 10\tabcolsep) * \real{0.1588}}
  >{\raggedright\arraybackslash}p{(\columnwidth - 10\tabcolsep) * \real{0.1588}}
  >{\raggedright\arraybackslash}p{(\columnwidth - 10\tabcolsep) * \real{0.1588}}
  >{\raggedright\arraybackslash}p{(\columnwidth - 10\tabcolsep) * \real{0.1765}}@{}}
\toprule\noalign{}
\begin{minipage}[b]{\linewidth}\raggedright
Method
\end{minipage} & \begin{minipage}[b]{\linewidth}\raggedright
Primary analysis
\end{minipage} & \begin{minipage}[b]{\linewidth}\raggedright
Small learning update
\end{minipage} & \begin{minipage}[b]{\linewidth}\raggedright
Moderate learning update
\end{minipage} & \begin{minipage}[b]{\linewidth}\raggedright
Large learning update
\end{minipage} & \begin{minipage}[b]{\linewidth}\raggedright
Route-direction update
\end{minipage} \\
\midrule\noalign{}
\endhead
\bottomrule\noalign{}
\endlastfoot
Direct reuse & 0.008471 & 0.008571 & 0.008371 & 0.008724 & 0.014458 \\
Fresh uniform & 0.067821 & 0.068636 & 0.067007 & 0.061789 & 0.069343 \\
Fresh adaptive & 0.065306 & 0.066243 & 0.064370 & 0.059589 & 0.066096 \\
Gap-based refresh & 0.006494 & 0.006602 & 0.006385 & 0.006808 &
0.011174 \\
Occupancy-sensitive refresh & 0.006525 & 0.006621 & 0.006429 & 0.006789
& 0.011080 \\
Transport uniform & 0.007236 & 0.007241 & 0.007232 & 0.009175 &
0.008080 \\
Transport adaptive & 0.006879 & 0.006834 & 0.006923 & 0.008982 &
0.007897 \\
DR uniform & 0.008916 & 0.008917 & 0.008915 & 0.010164 & 0.010324 \\
DR adaptive & 0.008462 & 0.008339 & 0.008585 & 0.009919 & 0.009753 \\
WIS adaptive & 0.006866 & 0.006727 & 0.007005 & 0.008387 & 0.008131 \\
\end{longtable}

Table A6. AUC by subgroup under small and moderate updates. Subgroups
overlap and should not be summed; each slice is descriptive.

\begin{longtable}[]{@{}
  >{\raggedright\arraybackslash}p{(\columnwidth - 10\tabcolsep) * \real{0.1530}}
  >{\raggedright\arraybackslash}p{(\columnwidth - 10\tabcolsep) * \real{0.0941}}
  >{\raggedright\arraybackslash}p{(\columnwidth - 10\tabcolsep) * \real{0.1882}}
  >{\raggedright\arraybackslash}p{(\columnwidth - 10\tabcolsep) * \real{0.1882}}
  >{\raggedright\arraybackslash}p{(\columnwidth - 10\tabcolsep) * \real{0.1882}}
  >{\raggedright\arraybackslash}p{(\columnwidth - 10\tabcolsep) * \real{0.1882}}@{}}
\toprule\noalign{}
\begin{minipage}[b]{\linewidth}\raggedright
Population
\end{minipage} & \begin{minipage}[b]{\linewidth}\raggedright
Tasks
\end{minipage} & \begin{minipage}[b]{\linewidth}\raggedright
Gap-based refresh
\end{minipage} & \begin{minipage}[b]{\linewidth}\raggedright
Occupancy-sensitive refresh
\end{minipage} & \begin{minipage}[b]{\linewidth}\raggedright
Transport adaptive
\end{minipage} & \begin{minipage}[b]{\linewidth}\raggedright
DR adaptive
\end{minipage} \\
\midrule\noalign{}
\endhead
\bottomrule\noalign{}
\endlastfoot
All & 288 & 0.006067 & 0.006105 & 0.006469 & 0.007991 \\
Competing routes & 216 & 0.006494 & 0.006525 & 0.006879 & 0.008462 \\
Serial controls & 72 & 0.004789 & 0.004845 & 0.005240 & 0.006577 \\
Horizon 6 & 96 & 0.006258 & 0.006280 & 0.006434 & 0.006647 \\
Horizon 9 & 96 & 0.005424 & 0.005414 & 0.005989 & 0.008592 \\
Horizon 12 & 96 & 0.006520 & 0.006620 & 0.006983 & 0.008733 \\
Build & 96 & 0.006390 & 0.006418 & 0.006794 & 0.008248 \\
Data & 96 & 0.005260 & 0.005297 & 0.005660 & 0.007600 \\
Publish & 96 & 0.006552 & 0.006600 & 0.006953 & 0.008124 \\
\end{longtable}

Table A7. Exact error decomposition on the full confirmation set. Mean
old value is 0.6823 in every row; the exact first-order remainder is
diagnostic only.

\begin{longtable}[]{@{}
  >{\raggedright\arraybackslash}p{(\columnwidth - 8\tabcolsep) * \real{0.1412}}
  >{\raggedright\arraybackslash}p{(\columnwidth - 8\tabcolsep) * \real{0.2059}}
  >{\raggedright\arraybackslash}p{(\columnwidth - 8\tabcolsep) * \real{0.2059}}
  >{\raggedright\arraybackslash}p{(\columnwidth - 8\tabcolsep) * \real{0.2235}}
  >{\raggedright\arraybackslash}p{(\columnwidth - 8\tabcolsep) * \real{0.2235}}@{}}
\toprule\noalign{}
\begin{minipage}[b]{\linewidth}\raggedright
Update
\end{minipage} & \begin{minipage}[b]{\linewidth}\raggedright
Drift MSE
\end{minipage} & \begin{minipage}[b]{\linewidth}\raggedright
Remainder MSE
\end{minipage} & \begin{minipage}[b]{\linewidth}\raggedright
$\epsilon$-relevant reversal rate
\end{minipage} & \begin{minipage}[b]{\linewidth}\raggedright
New mean value
\end{minipage} \\
\midrule\noalign{}
\endhead
\bottomrule\noalign{}
\endlastfoot
Small learning update & 8.468e-05 & 3.317e-08 & 0.000\% & 0.6890 \\
Moderate learning update & 4.203e-04 & 8.169e-07 & 0.116\% & 0.6973 \\
Large learning update & 3.096e-03 & 4.660e-05 & 2.083\% & 0.7238 \\
Route-direction update & 6.257e-03 & 1.233e-04 & 4.398\% & 0.6876 \\
\end{longtable}

In the error decomposition, mean success averages the true values of all
three root actions. It is not task success after selecting the optimal
root action. No task has all three candidate values equal to zero, but
this does not imply that individual routes are free of failure risk.

\subsection{Quality Crossing and New-Execution
Budgets}\label{f.4-quality-crossing-and-new-execution-budgets}

The prespecified quality target requires both mean regret \textless=
0.02 and probability of choosing an action with regret \textgreater{}
0.02 \textless= 0.05. The quality crossing is the earliest point on the
budget grid at which the target is met and remains met at every larger
budget. We do not interpolate, and cases that never meet the target
remain marked as not reached.

Table A8. Earliest allowed budget at which competing routes sustain the
quality target. Unreached targets are retained.

\begin{longtable}[]{@{}
  >{\raggedright\arraybackslash}p{(\columnwidth - 12\tabcolsep) * \real{0.1589}}
  >{\raggedright\arraybackslash}p{(\columnwidth - 12\tabcolsep) * \real{0.0765}}
  >{\raggedright\arraybackslash}p{(\columnwidth - 12\tabcolsep) * \real{0.1529}}
  >{\raggedright\arraybackslash}p{(\columnwidth - 12\tabcolsep) * \real{0.1529}}
  >{\raggedright\arraybackslash}p{(\columnwidth - 12\tabcolsep) * \real{0.1529}}
  >{\raggedright\arraybackslash}p{(\columnwidth - 12\tabcolsep) * \real{0.1529}}
  >{\raggedright\arraybackslash}p{(\columnwidth - 12\tabcolsep) * \real{0.1529}}@{}}
\toprule\noalign{}
\begin{minipage}[b]{\linewidth}\raggedright
Update
\end{minipage} & \begin{minipage}[b]{\linewidth}\raggedright
Old N
\end{minipage} & \begin{minipage}[b]{\linewidth}\raggedright
Gap
\end{minipage} & \begin{minipage}[b]{\linewidth}\raggedright
Occupancy
\end{minipage} & \begin{minipage}[b]{\linewidth}\raggedright
Transport
\end{minipage} & \begin{minipage}[b]{\linewidth}\raggedright
DR
\end{minipage} & \begin{minipage}[b]{\linewidth}\raggedright
Fresh uniform
\end{minipage} \\
\midrule\noalign{}
\endhead
\bottomrule\noalign{}
\endlastfoot
Small learning update & 16 & Not reached & Not reached & Not reached &
Not reached & Not reached \\
Small learning update & 64 & 768 & 768 & 768 & 1536 & Not reached \\
Small learning update & 256 & 0 & 0 & 0 & 0 & Not reached \\
Moderate learning update & 16 & Not reached & Not reached & Not reached
& Not reached & Not reached \\
Moderate learning update & 64 & 1536 & 1536 & 1536 & 1536 & Not
reached \\
Moderate learning update & 256 & 0 & 0 & 0 & 0 & Not reached \\
Large learning update & 16 & Not reached & Not reached & Not reached &
Not reached & Not reached \\
Large learning update & 64 & 1536 & 1536 & Not reached & Not reached &
Not reached \\
Large learning update & 256 & 768 & 768 & 0 & 0 & Not reached \\
Route-direction update & 16 & Not reached & 1536 & Not reached & Not
reached & Not reached \\
Route-direction update & 64 & Not reached & 1536 & 1536 & 1536 & Not
reached \\
Route-direction update & 256 & Not reached & Not reached & 0 & 0 & Not
reached \\
\end{longtable}

These budgets are discrete point estimates, not savings rates with
confidence guarantees. In particular, an AUC improvement does not imply
at least 25\% lower execution. That magnitude was only the target effect
size in the original protocol.

\subsection{Independent L4 Gate
Test}\label{f.5-independent-l4-gate-test}

Table A10 summarizes confirmatory comparisons in the primary population,
which equally weights ordinary and branch-selective small or moderate
updates under the prespecified protocol. The upper bound of the regret
difference between DSC-Gate and Gap-Gate is +0.00018, below the 0.0005
noninferiority margin. Ratios of new tool steps relative to Gap-Gate,
WIS-Gate, and DR-Gate are 0.606, 0.549, and 0.513, with all three
simultaneous intervals below 1. Regret is also lower than under both
WIS-Gate and DR-Gate.

Table A10. Confirmatory gate comparisons in the L4 primary population. A
cost ratio below 1 indicates fewer new tool steps under DSC-Gate.

\begin{longtable}[]{@{}
  >{\raggedright\arraybackslash}p{(\columnwidth - 8\tabcolsep) * \real{0.1601}}
  >{\raggedright\arraybackslash}p{(\columnwidth - 8\tabcolsep) * \real{0.2094}}
  >{\raggedright\arraybackslash}p{(\columnwidth - 8\tabcolsep) * \real{0.2261}}
  >{\raggedright\arraybackslash}p{(\columnwidth - 8\tabcolsep) * \real{0.1798}}
  >{\raggedright\arraybackslash}p{(\columnwidth - 8\tabcolsep) * \real{0.2245}}@{}}
\toprule\noalign{}
\begin{minipage}[b]{\linewidth}\raggedright
Comparison
\end{minipage} & \begin{minipage}[b]{\linewidth}\raggedright
Regret difference DSC - baseline
\end{minipage} & \begin{minipage}[b]{\linewidth}\raggedright
95\% interval
\end{minipage} & \begin{minipage}[b]{\linewidth}\raggedright
Tool-step ratio
\end{minipage} & \begin{minipage}[b]{\linewidth}\raggedright
Simultaneous 95\% interval
\end{minipage} \\
\midrule\noalign{}
\endhead
\bottomrule\noalign{}
\endlastfoot
vs Gap-Gate & +0.00004 & {[}-0.00010,+0.00018{]} & 0.606 &
{[}0.550,0.668{]} \\
vs WIS-Gate & -0.00061 & {[}-0.00089,-0.00034{]} & 0.549 &
{[}0.500,0.604{]} \\
vs DR-Gate & -0.00139 & {[}-0.00179,-0.00100{]} & 0.513 &
{[}0.462,0.568{]} \\
\end{longtable}

Table A11 shows how the gate path changes with the update condition.
Under ordinary updates, DSC-Gate resolves 55.4\% of comparisons by
reuse, 22.8\% by transport, and 21.8\% by refresh, using 151 new steps
on average. Under branch-selective updates, the proportions are 31.8\%,
12.3\%, and 55.9\%, with 421 new steps. Equal weighting of the two
update classes gives the main-text refresh rate of 38.9\% and mean cost
of 286 new tool steps.

Table A11. Conditional gate paths under DSC-Gate.

\begin{longtable}[]{@{}
  >{\raggedright\arraybackslash}p{(\columnwidth - 10\tabcolsep) * \real{0.2279}}
  >{\raggedright\arraybackslash}p{(\columnwidth - 10\tabcolsep) * \real{0.1324}}
  >{\raggedright\arraybackslash}p{(\columnwidth - 10\tabcolsep) * \real{0.1471}}
  >{\raggedright\arraybackslash}p{(\columnwidth - 10\tabcolsep) * \real{0.1324}}
  >{\raggedright\arraybackslash}p{(\columnwidth - 10\tabcolsep) * \real{0.1985}}
  >{\raggedright\arraybackslash}p{(\columnwidth - 10\tabcolsep) * \real{0.1618}}@{}}
\toprule\noalign{}
\begin{minipage}[b]{\linewidth}\raggedright
Update condition
\end{minipage} & \begin{minipage}[b]{\linewidth}\raggedright
Reuse
\end{minipage} & \begin{minipage}[b]{\linewidth}\raggedright
Transport
\end{minipage} & \begin{minipage}[b]{\linewidth}\raggedright
Refresh
\end{minipage} & \begin{minipage}[b]{\linewidth}\raggedright
Mean new tool steps
\end{minipage} & \begin{minipage}[b]{\linewidth}\raggedright
Mean regret
\end{minipage} \\
\midrule\noalign{}
\endhead
\bottomrule\noalign{}
\endlastfoot
Ordinary small or moderate & 55.4\% & 22.8\% & 21.8\% & 151 & 0.00491 \\
Branch-selective small or moderate & 31.8\% & 12.3\% & 55.9\% & 421 &
0.00855 \\
Equal-weight aggregate & 43.6\% & 17.6\% & 38.9\% & 286 & 0.00673 \\
\end{longtable}

\section{Cost Accounting and Reproducibility
Materials}\label{g-cost-accounting-and-reproducibility-materials}

Table A9. L2-L3 cost accounting, aggregated across tasks. Shared data
are counted once.

\begin{longtable}[]{@{}
  >{\raggedright\arraybackslash}p{(\columnwidth - 4\tabcolsep) * \real{0.4705}}
  >{\raggedright\arraybackslash}p{(\columnwidth - 4\tabcolsep) * \real{0.2941}}
  >{\raggedright\arraybackslash}p{(\columnwidth - 4\tabcolsep) * \real{0.2353}}@{}}
\toprule\noalign{}
\begin{minipage}[b]{\linewidth}\raggedright
Item
\end{minipage} & \begin{minipage}[b]{\linewidth}\raggedright
Count
\end{minipage} & \begin{minipage}[b]{\linewidth}\raggedright
Unit
\end{minipage} \\
\midrule\noalign{}
\endhead
\bottomrule\noalign{}
\endlastfoot
Policy-update trajectories & 1,229,204 & tool steps \\
Largest old-trajectory set & 12,293,461 & tool steps \\
Development and calibration references & 15,741,039 & tool steps \\
Target-pool simulation & 41,824,782 & tool steps \\
Model prompts & 11,454 & prompts \\
\end{longtable}

Old-data budgets are nested prefixes of the same largest trajectory set.
Estimators share old data, policy updates, and the target pool, so costs
must not be counted again for each budget or method. Development and
calibration reference trajectories must be counted in full. Amortized
once over the 288 confirmation tasks, they cost 54,656.39 tool steps per
task. The 41,824,782 target-pool steps belong to counterfactual
evaluation infrastructure and do not represent the online budget paid by
any one learner. Actual new execution is recorded separately in the
per-budget result tables.

Suppose the old data have already been paid for, reference quantities
can be reused for later in-distribution decisions, and a chosen quality
target saves s \textgreater{} 0 new steps per decision. The static
number of decisions required to amortize the reference cost is then
ceil(15,741,039/s). This scenario depends on transferability and the
reliability of the quality crossing; it is not an overall cost saving
already demonstrated by the experiment.

The L1 materials include the locked protocol, task configurations,
task-level summaries, compressed action-pair results, and confirmation
tests. The L2-L3 lightweight materials include frozen configurations,
model and dependency records, simulator and
estimation-allocation-aggregation code, and the complete tensor of
confirmation metrics. The metric tensor is indexed by task, update,
seed, old-data budget, method, new-execution budget, and metric, with
shape 288 x 4 x 5 x 3 x 10 x 8 x 5. The L2 tensor has shape 288 x 4 x 5
x 3 x 4 x 6. We also save exact old and new values and exact directional
derivatives, enabling aggregation checks and regeneration of diagnostic
plots. The lightweight package omits complete raw trajectories;
trajectory-level reruns require either the full raw data or new
simulation under the locked configuration. Figures 1-4 and Figure A1 use
every corresponding record rather than selected representative
trajectories. L4 uses protocol\_gate.json, frozen before test results
were read. It prespecifies the independent test tasks, the primary
population equally weighting ordinary and branch-selective small or
moderate updates, the three old-data budgets, five evaluation seeds,
regret noninferiority margin of 0.0005, 1,536-tool-step cap, and
stratified paired-bootstrap procedure. L4 artifacts retain task-level
stopping states, actual new tool steps, final choices, pair-level gate
states, and bootstrap indices for auditing the main text and Tables
A10-A11. L4 tasks are disjoint from all L1-L3 development, calibration,
and confirmation tasks. After freezing, gate rules, the primary
population, and statistical thresholds were not adjusted in response to
test regret, refresh cost, or between-method differences.

\end{document}